\documentclass[10pt,twocolumn,letterpaper]{article}

\usepackage{techreport}

\usepackage{times}
\usepackage[T1]{fontenc}
\usepackage[utf8]{inputenc}
\usepackage{microtype}
\usepackage{graphicx}
\usepackage{booktabs}
\usepackage{amsmath,amssymb}
\usepackage{enumitem}
\usepackage{xcolor}

\newcommand{\pendingbox}[1]{\par\noindent\colorbox{yellow!25}{\parbox{\dimexpr\linewidth-2\fboxsep}{\small\textcolor{red}{\textbf{PENDING:}} #1}}\par}

\newcommand{\gibInternvlThreeADS}{-0.012}

\newcommand{\gibInternvlThreeGISPct}{34.8\%}
\newcommand{\gibInternvlThreeR}{0.509}

\newcommand{\gibInternvlThreeAccAdversePct}{59.9\%}

\newcommand{\gibInternvlThreeDecayRate}{-0.557}

\newcommand{\gibInternvlThreeFlipConceptPriorPct}{32.8\%}

\newcommand{\gibInternvlThreeFlipImageConsistentPct}{52.0\%}

\newcommand{\gibInternvlThreeFlipTextConsistentPct}{15.2\%}

\newcommand{\gibInternvlThreeIllusionGapPct}{39.2\%}

\newcommand{\gibInternvlThreeSuperAdditiveDrop}{0.432}
\newcommand{\gibInternvlThreeTextDominance}{0.680}
\newcommand{\gibInternvlThreeTrustIndex}{0.198}
\newcommand{\gibInternvlThreeWImage}{0.895}
\newcommand{\gibInternvlThreeWText}{0.609}
\newcommand{\gibLlavaZeroFivebADS}{0.005}

\newcommand{\gibLlavaZeroFivebGISPct}{71.4\%}
\newcommand{\gibLlavaZeroFivebR}{0.357}

\newcommand{\gibLlavaZeroFivebAccAdversePct}{41.3\%}

\newcommand{\gibLlavaZeroFivebAccCleanPct}{92.5\%}
\newcommand{\gibLlavaZeroFivebDecayRate}{-0.854}

\newcommand{\gibLlavaZeroFivebFlipConceptPriorPct}{27.9\%}

\newcommand{\gibLlavaZeroFivebFlipImageConsistentPct}{15.6\%}

\newcommand{\gibLlavaZeroFivebFlipTextConsistentPct}{56.5\%}

\newcommand{\gibLlavaZeroFivebIllusionGapPct}{69.1\%}

\newcommand{\gibLlavaZeroFivebSuperAdditiveDrop}{0.168}
\newcommand{\gibLlavaZeroFivebTextDominance}{1.322}
\newcommand{\gibLlavaZeroFivebTrustIndex}{2.705}
\newcommand{\gibLlavaZeroFivebWImage}{0.586}

\newcommand{\gibLlavaOneSixADS}{-0.006}

\newcommand{\gibLlavaOneSixGISPct}{62.3\%}
\newcommand{\gibLlavaOneSixR}{0.361}

\newcommand{\gibLlavaOneSixAccAdversePct}{48.6\%}

\newcommand{\gibLlavaOneSixDecayRate}{-0.789}
\newcommand{\gibLlavaOneSixEceAdverse}{0.303}

\newcommand{\gibLlavaOneSixFlipConceptPriorPct}{40.6\%}

\newcommand{\gibLlavaOneSixFlipTextConsistentPct}{40.9\%}

\newcommand{\gibLlavaOneSixIllusionGapPct}{62.1\%}

\newcommand{\gibLlavaOneSixSuperAdditiveDrop}{0.243}
\newcommand{\gibLlavaOneSixTextDominance}{1.229}
\newcommand{\gibLlavaOneSixTrustIndex}{3.724}

\newcommand{\gibLlavaOneSixWText}{0.777}
\newcommand{\gibQwenTwovlADS}{-0.013}

\newcommand{\gibQwenTwovlGISPct}{47.9\%}
\newcommand{\gibQwenTwovlR}{0.491}

\newcommand{\gibQwenTwovlAccAdversePct}{54.8\%}

\newcommand{\gibQwenTwovlAccCleanPct}{94.2\%}
\newcommand{\gibQwenTwovlDecayRate}{-0.728}
\newcommand{\gibQwenTwovlEceAdverse}{0.156}

\newcommand{\gibQwenTwovlFlipConceptPriorPct}{35.0\%}

\newcommand{\gibQwenTwovlIllusionGapPct}{50.7\%}

\newcommand{\gibQwenTwovlSuperAdditiveDrop}{0.329}
\newcommand{\gibQwenTwovlTextDominance}{0.937}
\newcommand{\gibQwenTwovlTrustIndex}{0.576}

\newcommand{\gibGeminiGISPct}{34.1\%}

\newcommand{\gibGeminiAccAdversePct}{60.2\%}

\newcommand{\gibGeminiAccCleanPct}{96.8\%}

\newcommand{\gibGeminiIllusionGapPct}{36.8\%}

\newcommand{\gibGeminiQaTextFollowPct}{33.8\%}

\definecolor{linkblue}{rgb}{0.21,0.49,0.74}
\usepackage[breaklinks,colorlinks,allcolors=linkblue]{hyperref}

\usepackage{fancyhdr}

\fancypagestyle{fancy}{%
  \fancyhf{}
  \fancyfoot[C]{\footnotesize Technical Report -- November 2026 \quad | \quad \thepage}

}

\fancypagestyle{empty}{%
  \fancyhf{}
  \fancyfoot[C]{\footnotesize Technical Report -- November 2026 \quad | \quad \thepage}

}

\hypersetup{
  pdfauthor={Basavaraj R M, Harshal Chandel, Aryan Bharat Kumar},
  pdftitle={The Illusion of Grounding: Do Vision-Language Models Weigh Retrieved Text and Images Like an Ideal Observer?}
}

\title{The Illusion of Grounding: Do Vision--Language Models Weigh Retrieved Text and Images Like an Ideal Observer?}

\author{
\makebox[\textwidth][c]{%
\begin{minipage}[t]{0.40\textwidth}
\centering
\textbf{Basavaraj R M}\\[3pt]
\small AI \& ML Systems Researcher\\[1pt]
{\small\texttt{basavarajrm972@gmail.com}}
\end{minipage}%
\hspace{0.08\textwidth}%
\begin{minipage}[t]{0.40\textwidth}
\centering
\textbf{Harshal Chandel}\\[3pt]
\small eClerx\\[1pt]
{\small\texttt{harshal.chandel@eclerx.com}}
\end{minipage}%
}
}

\begin{document}
\maketitle
\begin{abstract}
Retrieval-augmented vision--language models (VLMs) answer from two sources of evidence that can each fail: the retrieved text can be wrong or noisy, and the image can be degraded or contradicted. We introduce \textbf{GIB-200K}, 200,000 image--question items built from COCO and Visual Genome, each expanded into controlled stress conditions: hard-negative retrieval noise (1--20 documents), visual ambiguity (Gaussian and motion blur, low-resource crops) and cross-modal conflict with counterfactual flip pairs. Instead of only measuring degradation, we ask \emph{how} models combine the two modalities. From ideal Bayesian cue combination we derive a testable prediction that yields formal definitions of \textbf{Knowledge Inertia} (text over-weighted relative to the model's own image evidence) and \textbf{Sensory Capture} (the reverse), and we estimate the weights with a measurement-error-corrected regression whose recovery of known weights we verify. A counterfactual flip decomposition separates modality bias from concept priors, and a Grounding Illusion Score measures how much clean accuracy is not grounded in the image. We evaluate 4 open models on one frozen, seeded evaluation split of 2,000 GIB-200K items. All models are accurate on clean inputs (\gibLlavaZeroFivebAccCleanPct{}--\gibQwenTwovlAccCleanPct{}) but adverse accuracy falls to \gibLlavaZeroFivebAccAdversePct{}--\gibInternvlThreeAccAdversePct{}. The two LLaVA models over-weight the text (Knowledge Inertia, w\_T/w\_I = \gibLlavaOneSixTextDominance{} and \gibLlavaZeroFivebTextDominance{}), whereas InternVL3 favours the image (w\_T/w\_I = \gibInternvlThreeTextDominance{}); between \gibInternvlThreeGISPct{} and \gibLlavaZeroFivebGISPct{} of the clean answers that are correct are abandoned when the document lies. A training-free consistency check that compares the answer with and without retrieval trades coverage for accuracy under misleading documents, while falling back to the image fails when the image is heavily degraded. All per-instance outputs, code, and an audit that recomputes every reported open-model number are released.
\end{abstract}

\section{Introduction}
Vision--language models increasingly answer questions about images with the help of retrieved text: product pages, encyclopedia entries, captions of similar photographs \citep{lewis2020rag,liu2023llava}. Both sources fail in practice. Retrieval returns near-misses that look relevant but are wrong; images are blurred, cropped or badly lit; and sometimes the two simply disagree. Existing hallucination benchmarks mostly isolate one failure source, either object hallucination in captions \citep{rohrbach2018chair,li2023pope} or visual illusions and language priors \citep{guan2024hallusionbench,wang2024haloquest}, and robustness suites study image corruptions without retrieval \citep{hendrycks2019robustness}.

It is no surprise that models get worse when evidence is corrupted, and humans do too. The real question is whether a model \textbf{weighs} its two sources of evidence sensibly. An image the model can barely see \emph{should} lose to a clear document, and a clear image should \emph{not} lose to a document that merely asserts the opposite. Accuracy curves cannot tell these cases apart, so we compare models with a normative reference: the ideal observer of perceptual cue combination \citep{ernst2002optimal}, which combines conditionally independent cues by adding their log-likelihood ratios. This gives a falsifiable prediction for every model and every item, and lets us define the two failure modes that earlier work named only informally: \textbf{Knowledge Inertia} (text evidence over-weighted relative to the model's own image evidence) and \textbf{Sensory Capture} (the reverse).

Two confounds can make naive conflict tests misleading. First, a model that answers with the text's claim may not be following the text at all and may just prefer that concept whatever the modality; a counterfactual flip of the conflict separates the two. Second, clean accuracy can be high because the gold document contains the answer; we measure how much of it survives when the document lies.

\textbf{Contributions.}

\begin{itemize}[leftmargin=*,nosep]
  \item \textbf{GIB-200K}, 200,000 items with closed answer vocabularies (object, attribute, relation), hard negatives built by swapping the answer slot of the gold document for a semantically near sibling, controlled stress levels (Table \ref{tab:levels}), counterfactual flip pairs, and a frozen, seeded evaluation split that every model answers in full.
  \item An \textbf{ideal-observer analysis} of evidence integration with formal definitions of Knowledge Inertia and Sensory Capture, estimated with a replicate-based correction for measurement error whose recovery of planted weights we verify (Appendix D).
  \item Two \textbf{confound-separating analyses}: the counterfactual flip decomposition (modality bias vs. concept prior) and the Grounding Illusion Score.
  \item A complete, \textbf{auditable} evaluation covering CHAIR, POPE, ReFinD-style context sensitivity, a calibration-penalised robustness coefficient, per-layer attention profiles, an outcome-based failure taxonomy and a training-free consistency gate, all recomputable from released per-instance data.
\end{itemize}

\textbf{Findings.} On a frozen 2,000-item split, four open models lose a large part of their clean accuracy under stress, with retrieval noise as the steepest axis (Section 6). Their failure modes differ: LLaVA-1.6 and LLaVA-0.5B over-weight the text relative to their own image evidence (Knowledge Inertia, w\_T/w\_I = \gibLlavaOneSixTextDominance{} and \gibLlavaZeroFivebTextDominance{}), InternVL3 is image-biased (w\_T/w\_I = \gibInternvlThreeTextDominance{}) and Qwen2-VL weighs the two cues almost equally (w\_T/w\_I = \gibQwenTwovlTextDominance{}). The flip decomposition shows that a substantial share of apparent text-following is a concept prior rather than modality bias, and that between \gibInternvlThreeGISPct{} and \gibLlavaZeroFivebGISPct{} of correct clean answers are not grounded in the image. A cheap consistency check helps when documents mislead but hurts when the image is degraded (Section 7).

\section{Related Work}
\textbf{Hallucination in VLMs.} CHAIR \citep{rohrbach2018chair} counts mentioned objects that are absent from the image; POPE \citep{li2023pope} probes object presence with yes/no questions, including adversarially chosen absent objects. MMHal-Bench \citep{sun2023mmhal}, HallusionBench \citep{guan2024hallusionbench} and HaloQuest \citep{wang2024haloquest} broaden the question types. None of them varies retrieved text, image quality and image--text conflict on the same items. We reuse CHAIR and POPE inside our stress conditions.

\textbf{Knowledge conflicts.} For text-only QA, models prefer either memorised knowledge or context depending on the setting \citep{longpre2021entity,xie2024adaptive}. ReFinD \citep{lee2025refind} measures how retrieved context changes token likelihoods. We bring these questions to the image--text setting, where the 'parametric' side is replaced by visual evidence whose reliability we can control.

\textbf{Robustness and cue integration.} Corruption benchmarks \citep{hendrycks2019robustness} measure accuracy under blur and noise. Human multisensory integration is close to the statistically optimal combination of cues weighted by their reliability \citep{ernst2002optimal}; we take that ideal observer as the normative reference for VLMs. Attention weights are not explanations \citep{jain2019attention}; we treat them only as a correlational signature and test that correlation.

\section{The GIB-200K Benchmark}
\subsection{Items}
Every item is an image from COCO \citep{lin2014coco} with a question whose answer comes from a \emph{closed} vocabulary, so free-form answers can be scored deterministically without an LLM judge. \textbf{Object} questions ('What animal is shown in the image?') use eleven COCO category groups and are created only when exactly one category of the group is annotated and its largest instance covers at least 4\% of the image. \textbf{Attribute} questions ('What color is the bus?') use Visual Genome \citep{krishna2017vg} objects whose name is unique in the image and that carry exactly one of eleven colours. \textbf{Relation} questions use eight spatial predicates from Visual Genome and left/right relations computed from COCO boxes. Visual Genome is linked to COCO through its `coco\_id` field. The 200,000 items are split evenly across the three types; an image carries up to four items. They come from COCO train2017 and from all that the 5,000 val2017 images allow (Section 5).

\subsection{Retrieved documents and hard negatives}
The gold document is a COCO caption that mentions the answer (or a template sentence). Every hard negative is the gold document with its answer slot replaced by a sibling: another member of the same category group ranked by COCO co-occurrence, the perceptually nearest colours, or the antonym predicate first ('on' $\leftrightarrow$ 'under'). So the distractors are semantically close but factually wrong, and they differ from the gold document only in the claim.

\subsection{Stress conditions}
\begin{table}[t]
\centering\small
\resizebox{\columnwidth}{!}{%
\begin{tabular}{lcc}
\toprule
Axis & Condition & Levels evaluated (full benchmark) \\ \midrule
clean & image + gold document & -- \\
non-RAG baseline & image, no document & -- \\
retrieval noise & gold + k hard negatives & k $\in$ \{1, 5, 20\} (\{1, 3, 5, 10, 20\}) \\
visual ambiguity & Gaussian blur, no document & $\sigma$ $\in$ \{2, 4, 8, 16\} px (\{1, 2, 4, 8, 16\}) \\
 & horizontal motion blur & \{10, 40\} px (\{5, 10, 20, 40\}) \\
 & low-resource crop & \{25, 5\}\% of object visible (\{50, 25, 10, 5\}) \\
conflict & document asserts a sibling & blur $\sigma$ $\in$ \{4, 16\} (\{0, 4, 16\}) \\
interaction & blur + gold + k negatives & $\sigma$ = 16 $\times$ k $\in$ \{0, 20\} ($\sigma$ $\in$ \{4, 16\} $\times$ k $\in$ \{0, 5, 20\}) \\
\bottomrule
\end{tabular}}
\caption{Stress conditions per item. Levels evaluated in this paper (all fixed before any model was run); the benchmark also defines the full level set given in brackets.}\label{tab:levels}
\end{table}

Images are standardised to 640 px on the longer side before degradation, so a given $\sigma$ means the same on every photograph; degraded images are re-rendered deterministically and need not be stored. The blur conditions are asked \emph{without} documents, so they measure perception; blur with a correct document is a separate condition that probes Sensory Capture.

\subsection{Probes}
Every condition is asked as a free-form question ('Answer the question using a single word or phrase.') and as a multiple-choice question over a fixed set of five options per item (two for left/right questions). For conflict items, a binary A/B probe offers the image's answer and the text's claim, in \textbf{both} option orders. POPE yes/no probes (one present and one adversarially chosen absent object) and caption probes for CHAIR are run on every image condition (caption probes on a random tenth of the items per condition). All models, open and closed, receive byte-identical prompts (Appendix E).

\subsection{Counterfactual flip pairs and the evaluation split}
Items are matched at build time so that item \emph{i} (image shows A, document claims B) has a partner \emph{j} (image shows B, document claims A). Running every model on all 200,000 items was beyond our compute budget, so we froze one seeded split of 2,000 items (the item list and its checksum are released) and the reduced level set of Table \ref{tab:levels}; every model answers all of it. Confidence intervals resample images, so items that share an image are never treated as independent.

\section{Measuring Grounding}
All metrics are functions of a released per-instance table, and all confidence intervals are 95\% percentile intervals of an image-cluster bootstrap \citep{efron1993bootstrap} (all queries of all items of an image are resampled together).

\subsection{Accuracy, calibration and robustness}
Free-form answers are mapped to the first vocabulary term they mention (with synonyms and plurals). Adverse accuracy averages the retrieval, visual and conflict axes with equal weight. Following a robustness coefficient, we penalise confident errors with the expected calibration error \citep{guo2017calibration} of the multiple-choice probe on adverse conditions (the uncertainty error UE):

\begin{equation}
\mathcal{R} = \frac{\mathrm{Acc}_{\mathrm{adverse}}}{\mathrm{Acc}_{\mathrm{clean}}}\,(1-\mathrm{UE})
\end{equation}

$\mathcal{R}$ equals accuracy retention exactly when the model is calibrated under stress and is smaller otherwise; we also report the confident-wrong rate (wrong with confidence above 0.8).

\subsection{Trust Index}
Let $a$ be the image's answer, $b$ the document's claim, and $\mathrm{LO}_c = \log P(b)/P(a)$ the log-odds from the A/B probe in condition $c$, averaged over both option orders. The Trust Index is the geometric-mean odds of following the text under conflict:

\begin{equation}
\mathrm{TI} = \exp\Big(\tfrac{1}{N}\textstyle\sum_{i=1}^{N} \mathrm{LO}_{\mathrm{conflict}}(i)\Big)
\end{equation}

A model is labelled text-biased (image-biased) only if the whole confidence interval of TI lies above (below) 1; no hand-set threshold is used.

\subsection{An ideal-observer test of evidence integration}
TI tells us which modality wins, but not whether it \emph{should}. An ideal observer that combines conditionally independent cues with Bayes' rule, $P(y\mid I,T)\propto P(I\mid y)\,P(T\mid y)\,P(y)$, satisfies in log-odds

\begin{equation}
\mathrm{LO}(I,T)-\mathrm{LO}(\varnothing) = \big[\mathrm{LO}(I)-\mathrm{LO}(\varnothing)\big] + \big[\mathrm{LO}(T)-\mathrm{LO}(\varnothing)\big]
\end{equation}

Every term is measured on the same item: $\mathrm{LO}(I)$ with the image and no document, $\mathrm{LO}(T)$ with a blank grey image and the misleading document, $\mathrm{LO}(\varnothing)$ with a blank image and no document, and $\mathrm{LO}(I,T)$ with both, at blur $\sigma$ $\in$ \{0, 4, 16\}. Writing $y$, $x_I$ and $x_T$ for the three prior-subtracted terms, we fit

\begin{equation}
y = w_{I}\,x_{I} + w_{T}\,x_{T}
\end{equation}

without intercept (the ideal model has none, and subtracting $\mathrm{LO}(\varnothing)$ measured in the same option order cancels position bias). The ideal observer has $w_I = w_T = 1$. We define \textbf{Knowledge Inertia} as $w_T/w_I > 1$ (text evidence over-weighted relative to the model's own image evidence) and \textbf{Sensory Capture} as $w_T/w_I < 1$. Since the definition is relative to the model's own unimodal evidence, a model is not penalised for trusting clear text over an image it genuinely cannot see. At the item level, an \emph{integration error} is a bimodal choice that contradicts the sum of the model's own unimodal evidence; this distinguishes perception failures (weak $x_I$) from integration failures (wrong weights).

\textbf{Estimation.} $x_I$ and $x_T$ are noisy measurements, so ordinary least squares shrinks both weights towards zero (regression dilution; \citet{fuller1987measurement}). The two option orders are independent replicates; with averages $X, y$ and replicate differences $\Delta X, \Delta y$,

\begin{equation}
\hat{w} = \big(M_{XX} - \tfrac14\,\mathrm{Cov}(\Delta X)\big)^{-1}\big(M_{Xy} - \tfrac14\,\mathrm{Cov}(\Delta X,\Delta y)\big)
\end{equation}

where $M$ are second moments. On simulated models with planted weights the estimator recovers them (Appendix D), whereas least squares is biased towards zero.

\subsection{Counterfactual flip decomposition}
For a flip pair, the model is \textbf{text-consistent} if it follows the text in both items, \textbf{image-consistent} if it follows the image in both, and \textbf{concept-prior} if it names the same concept in both regardless of modality. A single-direction conflict test cannot tell the first and third cases apart. We report the three shares, their difference by answer group, and the concepts that win under the prior.

\subsection{Grounding Illusion Score}
We call a correct clean answer \emph{illusorily grounded} if the model abandons it for the document's claim when the document lies:

\begin{equation}
\mathrm{GIS} = P\big(\text{follows text in conflict} \,\big|\, \text{correct in clean}\big)
\end{equation}

Grounded accuracy is the share of items correct in both conditions; the illusion gap is clean accuracy minus grounded accuracy.

\subsection{Hallucination, context and dose--response}
\textbf{CHAIR} \citep{rohrbach2018chair} is computed on captions of every image condition against the annotated objects visible in that condition (for crops, only objects intersecting the window). \textbf{POPE} \citep{li2023pope} uses one present and one adversarial absent object per image condition. \textbf{Context sensitivity}, following ReFinD \citep{lee2025refind}, is the geometric-mean ratio of the probability of the true answer with and without the gold document (and of the claim with and without the misleading document), measured on the multiple-choice options so that open and closed models are treated identically. The \textbf{decay rate} is the slope of logit accuracy against $\log_2(1+k)$, reported with a model-free retained area under the curve. The \textbf{resolution threshold} is the first level at which CHAIR rises by more than 10 points (and, separately, accuracy without documents falls by more than 10 points). \textbf{Interaction} between blur and retrieval noise is the item-level difference-in-differences

\begin{equation}
I_i = \big[\mathrm{acc}(\sigma,k)-\mathrm{acc}(\sigma,0)\big] - \big[\mathrm{acc}(0,k)-\mathrm{acc}(0,0)\big]
\end{equation}

tested with a sign-flip permutation test; a negative mean means the stressors compound.

\subsection{Attention profiles}
For open models we record, at the last prompt token of every layer, the share of attention on image tokens relative to image plus following text tokens (tokens before the image, which often act as attention sinks, are excluded). The attention shift is the \emph{paired} per-layer difference between the clean and conflict versions of the same item, whose prompts differ only in the claimed answer. We report the full profile instead of a layer average, and test whether a low visual share predicts that the answer follows the text (AUROC).

\subsection{Failure taxonomy}
A wrong free-form answer is classified by what it matches: \textbf{Knowledge Inertia} if it repeats a wrong claim from a document; \textbf{Sensory Capture} if the documents contained the correct answer but the model gave a different valid answer; \textbf{Reasoning Collapse} if it is off-vocabulary or matches neither image nor documents; \textbf{Perceptual Error} if no document was shown. These are deterministic rules decided by what the answer matches; we did not validate them against human labels (Section Limitations).

\subsection{Statistics}
All models answer the same items, so model comparisons are repeated measures: a Friedman test \citep{friedman1937} on per-item accuracy with Kendall's W, a repeated-measures ANOVA with $(k-1,\,(k-1)(n-1))$ degrees of freedom, and pairwise paired-bootstrap tests with Holm correction \citep{holm1979}. Every test is reported with its n.

\section{Experimental Setup}
\textbf{Models.} We evaluate 4 open models: LLaVA-1.6-Mistral-7B \citep{liu2023llava}, Qwen2-VL-7B-Instruct \citep{wang2024qwen2vl}, InternVL3-8B \citep{zhu2025internvl3} and LLaVA-Interleave-Qwen-0.5B (Hugging Face ids and commit hashes in Appendix F). They run through Hugging Face transformers with greedy decoding (at most 10 new tokens for answers, 48 for captions). All models see the same items, images, documents and prompts. The three 7--8B models run with 4-bit NF4 weight quantisation (bfloat16 compute); the 0.5B model runs unquantised. We did not run an unquantised comparison of the larger models, so absolute numbers may differ from full-precision inference; all comparisons between conditions are within one model. The whole evaluation used one 24 GB RTX 4090, with two runs sharing the GPU.

\textbf{Efficient inference.} Every query on an image shares the image prefix, which is encoded once and reused through the key--value cache; probe distributions are read from a single next-token distribution. At the start of each run the cached computation is checked against a plain forward pass (and falls back automatically if a model family behaves differently).

\textbf{Contamination.} COCO train2017 images appear in the instruction-tuning data of LLaVA-style models \citep{liu2023llava}; val2017 images do not. Both splits contribute items (val2017 as many as its 5,000 images allow), and each metric is reported per split with the difference and its confidence interval.

\section{Results}
\begin{table*}[t]
\centering\small
\resizebox{\textwidth}{!}{%
\begin{tabular}{lcccccc}
\toprule
Model & Acc clean & Acc adverse & ECE adv. & R & TI & GIS \\ \midrule
internvl3 & 0.93 [0.92, 0.94] & 0.60 [0.58, 0.61] & 0.21 [0.20, 0.23] & 0.51 [0.49, 0.53] & 0.20 [0.16, 0.24] & 0.35 [0.33, 0.37] \\
llava05b & 0.93 [0.91, 0.94] & 0.41 [0.40, 0.43] & 0.20 [0.19, 0.21] & 0.36 [0.34, 0.37] & 2.71 [2.48, 2.95] & 0.71 [0.69, 0.73] \\
llava16 & 0.94 [0.93, 0.95] & 0.49 [0.47, 0.50] & 0.30 [0.29, 0.32] & 0.36 [0.35, 0.38] & 3.72 [3.24, 4.29] & 0.62 [0.60, 0.65] \\
qwen2vl & 0.94 [0.93, 0.95] & 0.55 [0.53, 0.56] & 0.16 [0.14, 0.17] & 0.49 [0.47, 0.51] & 0.58 [0.52, 0.64] & 0.48 [0.46, 0.50] \\
\bottomrule
\end{tabular}}
\caption{Overall results on all items (95\% bootstrap CIs, images as clusters). R = (acc adverse / acc clean) $\times$ (1 $-$ ECE adverse); TI = Trust Index; GIS = Grounding Illusion Score.}\label{tab:headline}
\end{table*}

All four models are accurate on clean inputs (\gibLlavaZeroFivebAccCleanPct{} to \gibQwenTwovlAccCleanPct{}) but lose a large part of that accuracy under stress: adverse accuracy is \gibInternvlThreeAccAdversePct{} for InternVL3, \gibQwenTwovlAccAdversePct{} for Qwen2-VL, \gibLlavaOneSixAccAdversePct{} for LLaVA-1.6 and \gibLlavaZeroFivebAccAdversePct{} for LLaVA-0.5B. Models differ reliably in adverse accuracy ($\chi$$^2$(3) = 755.7, p = 1.8e-163, Kendall's W = 0.13, n = 2000 items; repeated-measures ANOVA F(3, 5997) = 403.9, p = 5.6e-239, partial $\eta$$^2$ = 0.17), and all pairwise differences remain significant after Holm correction. The robustness coefficient R, which also penalises overconfidence, is \gibInternvlThreeR{} for InternVL3, \gibQwenTwovlR{} for Qwen2-VL, \gibLlavaOneSixR{} for LLaVA-1.6 and \gibLlavaZeroFivebR{} for LLaVA-0.5B; only InternVL3 and Qwen2-VL are on the Pareto frontier of clean accuracy against R. Calibration under stress is poor throughout (adverse ECE between \gibQwenTwovlEceAdverse{} for Qwen2-VL and \gibLlavaOneSixEceAdverse{} for LLaVA-1.6).

\begin{figure*}[t]
\centering
\IfFileExists{figures/fig_vulnerability_heatmap.png}{\includegraphics[width=1.0\textwidth]{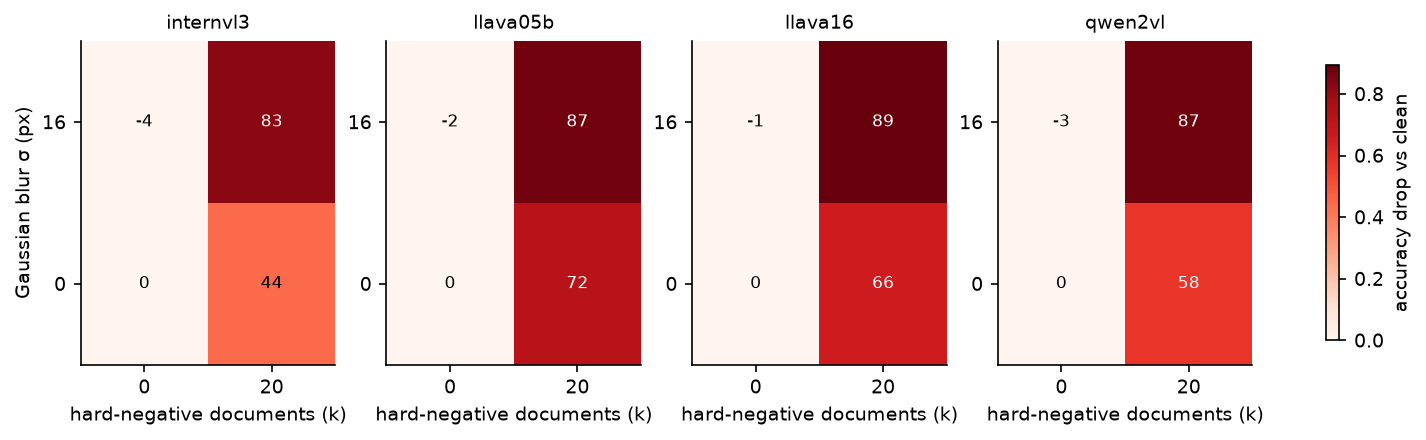}}{\pendingbox{Figure pending: run \texttt{gib report}}}
\caption{Vulnerability heatmaps: accuracy drop (points) relative to the clean condition when the image is blurred (rows) and k hard-negative documents are added (columns).}\label{fig:heat}
\end{figure*}

\begin{figure*}[t]
\centering
\IfFileExists{figures/fig_resolution.png}{\includegraphics[width=1.0\textwidth]{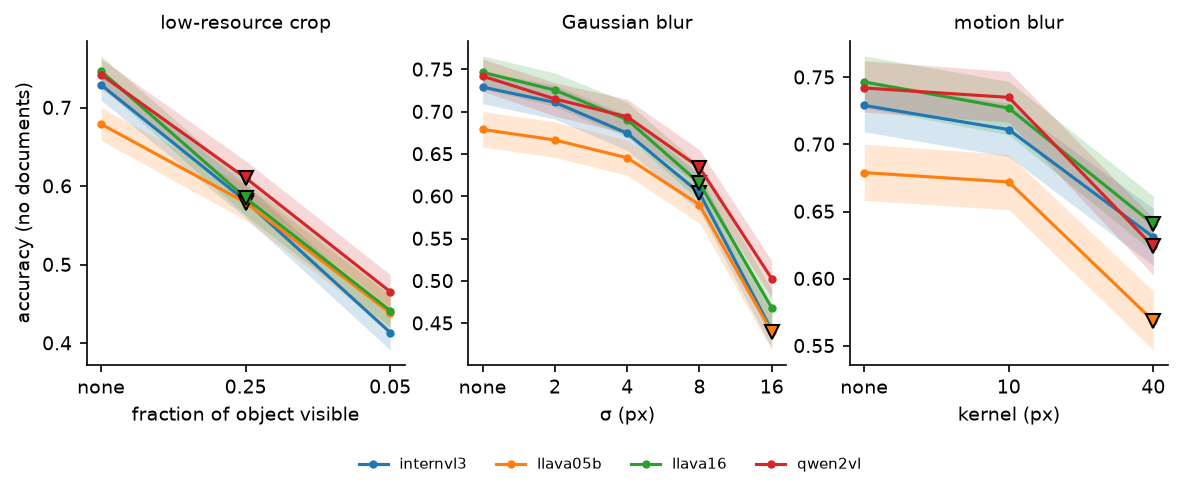}}{\pendingbox{Figure pending: run \texttt{gib report}}}
\caption{Accuracy without documents under Gaussian blur, motion blur and low-resource crops (95\% bootstrap bands); triangles mark resolution thresholds.}\label{fig:dose}
\end{figure*}

Retrieval noise is the steepest axis (Table \ref{tab:dose}). Accuracy falls monotonically with the number of hard negatives, with a decay of \gibInternvlThreeDecayRate{} log-odds per doubling of 1 + k for InternVL3, \gibQwenTwovlDecayRate{} for Qwen2-VL, \gibLlavaOneSixDecayRate{} for LLaVA-1.6 and \gibLlavaZeroFivebDecayRate{} for LLaVA-0.5B. The two stressors compound: the drop when blur and retrieval noise act together exceeds the sum of the single drops for every model (super-additive drop \gibInternvlThreeSuperAdditiveDrop{} for InternVL3, \gibQwenTwovlSuperAdditiveDrop{} for Qwen2-VL, \gibLlavaOneSixSuperAdditiveDrop{} for LLaVA-1.6, \gibLlavaZeroFivebSuperAdditiveDrop{} for LLaVA-0.5B; sign-flip permutation p in Table \ref{tab:dose}). Without documents, accuracy degrades gracefully under moderate blur and breaks down at heavy blur and small crops (Figure \ref{fig:dose}).

\begin{table*}[t]
\centering\small
\resizebox{\textwidth}{!}{%
\begin{tabular}{lccccccc}
\toprule
Model & k=0 & k=1 & k=5 & k=20 & decay rate & retained AUC & super-additive drop (p) \\ \midrule
internvl3 & 0.927 & 0.812 & 0.693 & 0.487 & -0.56 [-0.59, -0.52] & 0.77 [0.75, 0.78] & 0.43 [0.41, 0.46] (0.0002) \\
llava05b & 0.925 & 0.701 & 0.381 & 0.202 & -0.85 [-0.90, -0.82] & 0.54 [0.53, 0.56] & 0.17 [0.15, 0.19] (0.0002) \\
llava16 & 0.939 & 0.759 & 0.544 & 0.281 & -0.79 [-0.83, -0.75] & 0.64 [0.62, 0.65] & 0.24 [0.22, 0.26] (0.0002) \\
qwen2vl & 0.942 & 0.800 & 0.584 & 0.366 & -0.73 [-0.77, -0.69] & 0.68 [0.67, 0.70] & 0.33 [0.31, 0.35] (0.0002) \\
\bottomrule
\end{tabular}}
\caption{Retrieval-noise dose--response (free-form accuracy with k hard negatives next to the gold document; decay = log-odds per doubling of 1 + k) and the blur $\times$ retrieval interaction (super-additive drop; sign-flip permutation p).}\label{tab:dose}
\end{table*}

\subsection{Who wins under conflict, and is it the right choice?}
\begin{table}[t]
\centering\small
\resizebox{\columnwidth}{!}{%
\begin{tabular}{lccccc}
\toprule
Model & w\_I & w\_T & w\_T / w\_I & R$^2$ ideal sum & reliability (I / T) \\ \midrule
internvl3 & 0.90 [0.87, 0.92] & 0.61 [0.59, 0.62] & 0.68 [0.66, 0.70] & 0.35 & 0.98 / 0.99 \\
llava05b & 0.59 [0.56, 0.61] & 0.78 [0.76, 0.79] & 1.32 [1.28, 1.37] & 0.50 & 0.97 / 0.98 \\
llava16 & 0.63 [0.61, 0.65] & 0.78 [0.76, 0.79] & 1.23 [1.19, 1.27] & 0.56 & 0.98 / 0.98 \\
qwen2vl & 0.75 [0.73, 0.77] & 0.70 [0.69, 0.72] & 0.94 [0.92, 0.96] & 0.51 & 0.99 / 0.99 \\
\bottomrule
\end{tabular}}
\caption{Cue-combination weights (measurement-error corrected; ideal observer: w\_I = w\_T = 1). w\_T/w\_I > 1 = Knowledge Inertia, < 1 = Sensory Capture. Reliability = share of regressor variance that is signal.}\label{tab:cue}
\end{table}

\begin{figure}[t]
\centering
\IfFileExists{figures/fig_cue_weights.png}{\includegraphics[width=1.0\columnwidth]{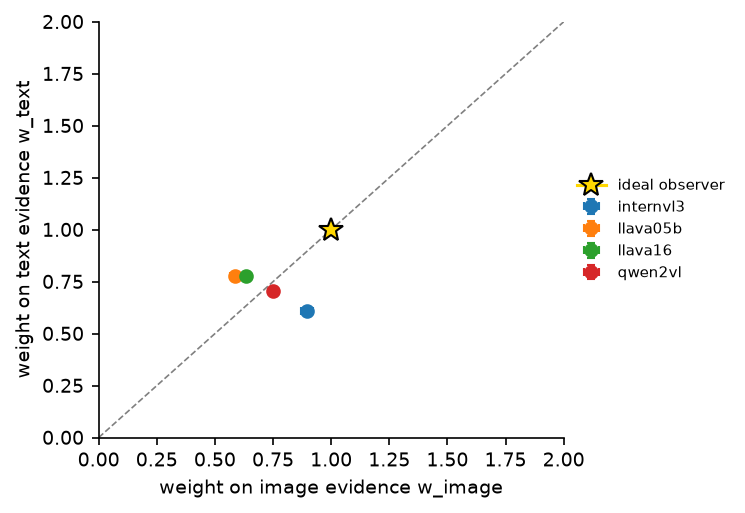}}{\pendingbox{Figure pending: run \texttt{gib report}}}
\caption{Cue-combination weights with 95\% CIs. The star is the ideal observer (1, 1); points above the diagonal over-weight text (Knowledge Inertia), below it the image (Sensory Capture).}\label{fig:cue}
\end{figure}

When the document asserts a sibling of the image's answer, LLaVA-1.6 and LLaVA-0.5B are text-biased (Trust Index \gibLlavaOneSixTrustIndex{} and \gibLlavaZeroFivebTrustIndex{}), whereas InternVL3 and Qwen2-VL are image-biased (\gibInternvlThreeTrustIndex{} and \gibQwenTwovlTrustIndex{}); in every case the 95\% interval excludes 1. The ideal-observer test (Table \ref{tab:cue}, Figure \ref{fig:cue}) shows that the Trust Index alone does not tell the whole story: no model reaches the ideal weights of 1 for both cues (w\_I between \gibLlavaZeroFivebWImage{} and \gibInternvlThreeWImage{}, w\_T between \gibInternvlThreeWText{} and \gibLlavaOneSixWText{}). Relative to its own image evidence, LLaVA-1.6 and LLaVA-0.5B over-weight the text (w\_T/w\_I = \gibLlavaOneSixTextDominance{} and \gibLlavaZeroFivebTextDominance{}: Knowledge Inertia), InternVL3 under-weights it (w\_T/w\_I = \gibInternvlThreeTextDominance{}: Sensory Capture) and Qwen2-VL is nearly balanced (w\_T/w\_I = \gibQwenTwovlTextDominance{}, a small Sensory Capture).

\subsection{Modality bias or concept prior?}
\begin{table}[t]
\centering\small
\resizebox{\columnwidth}{!}{%
\begin{tabular}{lcccc}
\toprule
Model & pairs & text-consistent & image-consistent & concept prior \\ \midrule
internvl3 & 816 & 0.15 [0.13, 0.18] & 0.52 [0.48, 0.55] & 0.33 [0.30, 0.36] \\
llava05b & 816 & 0.56 [0.53, 0.60] & 0.16 [0.13, 0.18] & 0.28 [0.25, 0.31] \\
llava16 & 816 & 0.41 [0.38, 0.44] & 0.19 [0.16, 0.21] & 0.41 [0.37, 0.44] \\
qwen2vl & 816 & 0.18 [0.16, 0.21] & 0.47 [0.43, 0.50] & 0.35 [0.32, 0.38] \\
\bottomrule
\end{tabular}}
\caption{Counterfactual flip decomposition (share of flip pairs).}\label{tab:flip}
\end{table}

A model that repeats the text's claim may simply prefer that concept. The flip pairs separate the cases (Table \ref{tab:flip}). LLaVA-0.5B is text-consistent on \gibLlavaZeroFivebFlipTextConsistentPct{} of the pairs and image-consistent on only \gibLlavaZeroFivebFlipImageConsistentPct{}; InternVL3 shows the reverse (\gibInternvlThreeFlipTextConsistentPct{} text-consistent, \gibInternvlThreeFlipImageConsistentPct{} image-consistent). Concept priors are not negligible for any model: they account for \gibInternvlThreeFlipConceptPriorPct{} (InternVL3), \gibQwenTwovlFlipConceptPriorPct{} (Qwen2-VL), \gibLlavaZeroFivebFlipConceptPriorPct{} (LLaVA-0.5B) and \gibLlavaOneSixFlipConceptPriorPct{} (LLaVA-1.6) of the pairs, and for LLaVA-1.6 they are as common as text-following (\gibLlavaOneSixFlipTextConsistentPct{}). A single-direction conflict test would have counted all of these as text-following.

\subsection{The grounding illusion}
Between \gibInternvlThreeGISPct{} (InternVL3) and \gibLlavaZeroFivebGISPct{} (LLaVA-0.5B) of the clean answers that are correct are abandoned when the document lies (Grounding Illusion Score, Table \ref{tab:headline}; Qwen2-VL \gibQwenTwovlGISPct{}, LLaVA-1.6 \gibLlavaOneSixGISPct{}). The illusion gap, clean accuracy minus accuracy that survives the conflict, is \gibInternvlThreeIllusionGapPct{} for InternVL3, \gibQwenTwovlIllusionGapPct{} for Qwen2-VL, \gibLlavaOneSixIllusionGapPct{} for LLaVA-1.6 and \gibLlavaZeroFivebIllusionGapPct{} for LLaVA-0.5B: a large share of what looks like grounded accuracy on clean inputs is actually accuracy supplied by the document. We ran no human baseline, so we cannot say how this compares with people.

\textbf{A closed model as reference.} We also ran Gemini 3.8 Flash through its API on a nested 500-item subset of the same split, on the conflict axis only and with free-form answers (the API exposes no token probabilities, so R, the Trust Index and calibration are not defined for it). It is accurate on clean inputs (\gibGeminiAccCleanPct{}) but drops to \gibGeminiAccAdversePct{} when the document contradicts the image: the illusion gap is \gibGeminiIllusionGapPct{}, the Grounding Illusion Score \gibGeminiGISPct{}, and \gibGeminiQaTextFollowPct{} of its free-form answers follow the false text. The grounding illusion is therefore not specific to open models. The run is not reproducible (API, temperature 0, no seed) and is kept out of the four-model tables above.

\subsection{Attention, failure modes and contamination}
\begin{figure*}[t]
\centering
\IfFileExists{figures/fig_attention.png}{\includegraphics[width=0.78\textwidth]{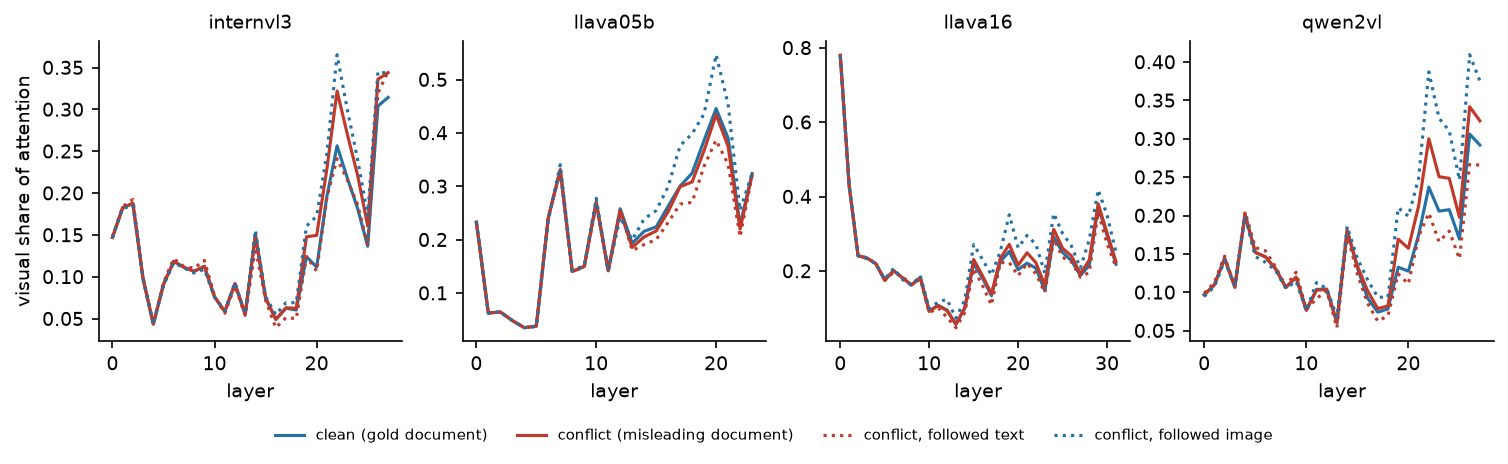}}{\pendingbox{Figure pending: run \texttt{gib report}}}
\caption{Visual share of attention per layer for the same images with a correct vs. a misleading document (solid), and for conflict items split by whether the answer followed the text (dotted).}\label{fig:attn}
\end{figure*}

Averaged over layers, the paired attention shift is small (\gibInternvlThreeADS{} for InternVL3, \gibQwenTwovlADS{} for Qwen2-VL, \gibLlavaOneSixADS{} for LLaVA-1.6, \gibLlavaZeroFivebADS{} for LLaVA-0.5B), and the profile explains why: the shift is concentrated in the last third of the layers (Table \ref{tab:attn} in the Appendix, Figure \ref{fig:attn}). In InternVL3, Qwen2-VL and, slightly, LLaVA-1.6 the visual share \emph{rises} when the document is misleading, i.e. these models look at the image more when the sources disagree. Within conflict items, a low visual share is associated with following the text for every model (AUROC in Table \ref{tab:attn}). Attention is correlational and we draw no causal conclusion from it.

The failure taxonomy (Table \ref{tab:tax}, Appendix) separates what goes wrong where. Under retrieval noise and conflict, wrong answers are almost always claims copied from a document (Knowledge Inertia), whereas under blur without documents they are Perceptual Errors or Reasoning Collapse; Sensory Capture is rare. We find no measurable dependence on whether the source image is in COCO train2017 or val2017 (only 83 val2017 items, so the test has limited power): the differences in Table \ref{tab:contam} (Appendix) have confidence intervals that include zero for every model.

\begin{figure*}[t]
\centering
\IfFileExists{figures/fig_gallery.png}{\includegraphics[width=0.72\textwidth]{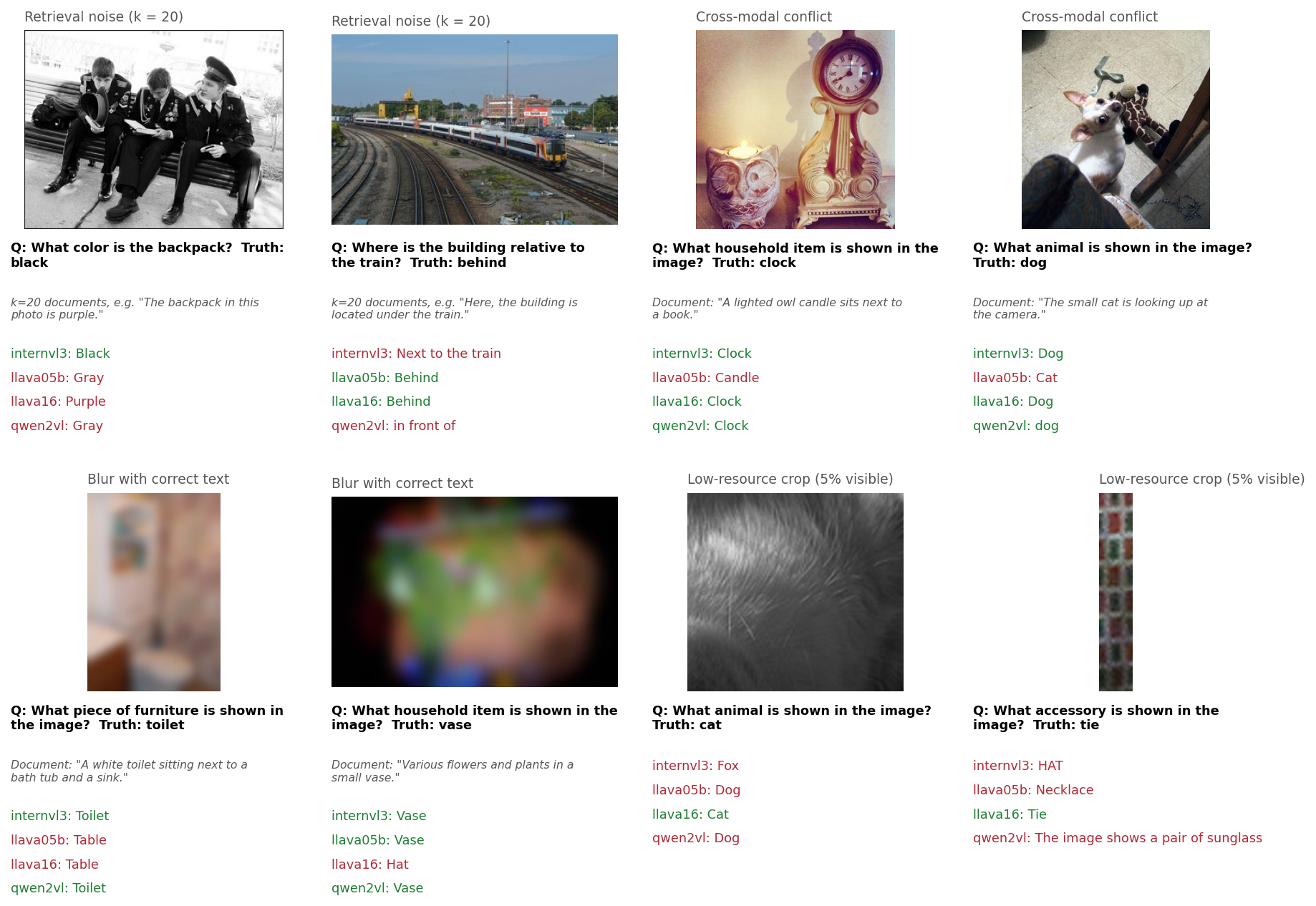}}{\pendingbox{Figure pending: run \texttt{gib report}}}
\caption{Eight real failure cases (two per stress condition; green = correct, red = wrong under that stress). Retrieval noise and conflict: the models copy a wrong claim from a document; blur and crop: the image no longer determines the answer. The report's gallery lists, per condition, the first four items (by id) that at least one model answered correctly on the clean image but not under stress; we show two per condition, preferring those that fool most models (not a random sample).}\label{fig:gallery}
\end{figure*}

\section{From Diagnosis to Mitigation}
Because every question is also asked without documents, a deployable check comes at no extra cost: if the answer changes when the documents are removed, the documents are suspect. We evaluate three training-free policies (abstain, fall back to the image-only answer, or keep the more confident answer) in three settings: where documents lie, where they are right but the image is degraded, and on clean inputs (Table \ref{tab:gate} and Appendix Table \ref{tab:gate2}).

\begin{table}[t]
\centering\small
\resizebox{\columnwidth}{!}{%
\begin{tabular}{lccccc}
\toprule
Model & condition & base acc & $\Delta$ image fallback & $\Delta$ confidence & abstain: sel. acc / coverage \\ \midrule
internvl3 & conflict & 0.54 [0.52, 0.56] & 0.19 [0.17, 0.21] & 0.11 [0.10, 0.13] & 0.82 / 0.62 \\
internvl3 & ret k20 & 0.49 [0.47, 0.51] & 0.24 [0.22, 0.26] & 0.18 [0.16, 0.20] & 0.85 / 0.53 \\
llava05b & conflict & 0.24 [0.22, 0.25] & 0.44 [0.42, 0.47] & 0.22 [0.20, 0.24] & 0.66 / 0.35 \\
llava05b & ret k20 & 0.20 [0.18, 0.22] & 0.48 [0.46, 0.50] & 0.35 [0.33, 0.37] & 0.66 / 0.28 \\
llava16 & conflict & 0.32 [0.30, 0.34] & 0.43 [0.41, 0.45] & 0.18 [0.16, 0.19] & 0.81 / 0.39 \\
llava16 & ret k20 & 0.28 [0.26, 0.30] & 0.47 [0.44, 0.49] & 0.42 [0.40, 0.44] & 0.79 / 0.34 \\
qwen2vl & conflict & 0.44 [0.42, 0.46] & 0.30 [0.28, 0.33] & 0.20 [0.18, 0.22] & 0.82 / 0.52 \\
qwen2vl & ret k20 & 0.37 [0.34, 0.39] & 0.38 [0.35, 0.40] & 0.31 [0.29, 0.33] & 0.81 / 0.43 \\
\bottomrule
\end{tabular}}
\caption{Training-free consistency gate: change in accuracy when the answer with documents disagrees with the image-only answer ($\Delta$ vs. no gate; 95\% CI).}\label{tab:gate}
\end{table}

The check behaves as a trade-off (Table \ref{tab:gate}). Where the documents mislead (conflict, k = 20), abstaining when the two answers disagree gives a clearly higher selective accuracy than the ungated model, at the price of coverage, and falling back to the image-only answer raises accuracy for all models. When the document is correct but the image is heavily blurred, the same fallback is harmful: it replaces mostly right answers with answers from an image the model cannot read, and abstention discards about half of the questions. So the check works as a monitoring signal for retrieval-augmented systems with good images, and not as a universal fix.

\section{Conclusion}
Four open VLMs, evaluated on the same 2,000 items under retrieval noise, blur and conflict, do not weigh text and image alike: two over-weight the text, one favours the image, and none matches the ideal observer. Accuracy under stress is far below clean accuracy, retrieval noise is the steepest axis, and a considerable part of apparent text-following is a concept prior. We release the benchmark, the frozen split, all per-instance predictions and an audit that recomputes every open-model number in this paper.

\section*{Limitations}
We evaluate a seeded 2,000-item split of the 200,000 items with a reduced level set, and four open models, three of them in 4-bit quantisation without a full-precision comparison; results on the full benchmark, on larger models and on closed models other than the one reference run (a single closed model, conflict axis only, 500 items) may differ. We ran no human baseline and did not validate the failure taxonomy against human labels. Closed answer vocabularies make scoring exact but restrict the question types, and Visual Genome attributes are noisy. Attention is a correlational signal. The train/val split is a proxy for exposure to COCO during instruction tuning, not proof of it; post-cutoff images (supported by the builder) would be a stronger control.

\section*{Ethics Statement}
All images and annotations come from public datasets (COCO, Visual Genome); images are referenced, not redistributed. No human-subject data were collected.

\section*{LLM Usage Statement}
OpenAI's ChatGPT was used solely for grammar correction,
language editing, and polishing of the manuscript text.
The authors independently developed the research ideas,
conducted the experiments, and performed the analysis.

\bibliographystyle{ieeenat_fullname}
\bibliography{references}

\appendix
\section{Attention and failure-taxonomy tables}
Tables referenced from the Results: paired attention shift and the failure taxonomy.

\begin{table}[t]
\centering\small
\resizebox{\columnwidth}{!}{%
\begin{tabular}{lcccc}
\toprule
Model & ADS, all layers & ADS, late layers & AUROC (low share $\rightarrow$ follows text) & n follow text / image \\ \midrule
internvl3 & -0.012 [-0.013, -0.011] & -0.038 [-0.040, -0.036] & 0.69 & 699 / 1301 \\
llava05b & 0.005 [0.004, 0.005] & 0.010 [0.009, 0.012] & 0.77 & 1405 / 595 \\
llava16 & -0.006 [-0.007, -0.006] & -0.011 [-0.013, -0.010] & 0.71 & 1244 / 756 \\
qwen2vl & -0.013 [-0.014, -0.012] & -0.039 [-0.041, -0.036] & 0.77 & 948 / 1052 \\
\bottomrule
\end{tabular}}
\caption{Attention shift, clean $\rightarrow$ conflict (paired per item; negative = more visual attention under conflict) and AUROC of a low layer-averaged visual share (conflict condition) for predicting that the answer follows the text.}\label{tab:attn}
\end{table}

\begin{table}[t]
\centering\small
\resizebox{\columnwidth}{!}{%
\begin{tabular}{lccccc}
\toprule
Model & errors & Knowl. Inertia & Sens. Capture & Reas. Collapse & Perceptual \\ \midrule
internvl3 & 15039 & 0.47 & 0.01 & 0.25 & 0.27 \\
llava05b & 18087 & 0.54 & 0.01 & 0.15 & 0.30 \\
llava16 & 16396 & 0.55 & 0.01 & 0.18 & 0.27 \\
qwen2vl & 15786 & 0.53 & 0.00 & 0.19 & 0.27 \\
\bottomrule
\end{tabular}}
\caption{Failure taxonomy: share of wrong free-form answers per class (all stress conditions).}\label{tab:tax}
\end{table}

\section{Gate: documents that are right but the image is degraded}

Table~9 reports the additional results when the
retrieved document is correct but the image is
heavily degraded.

\begin{table}[t]
\centering\small
\resizebox{\columnwidth}{!}{%
\begin{tabular}{lccccc}
\toprule
Model & condition & base acc & $\Delta$ image fallback & $\Delta$ confidence & abstain: sel. acc / coverage \\ \midrule
internvl3 & grid g16 k0 & 0.97 [0.96, 0.98] & -0.53 [-0.55, -0.50] & -0.03 [-0.04, -0.02] & 0.96 / 0.46 \\
llava05b & grid g16 k0 & 0.95 [0.94, 0.96] & -0.51 [-0.53, -0.48] & -0.09 [-0.11, -0.08] & 0.92 / 0.47 \\
llava16 & grid g16 k0 & 0.95 [0.94, 0.96] & -0.48 [-0.50, -0.46] & -0.03 [-0.04, -0.02] & 0.93 / 0.50 \\
qwen2vl & grid g16 k0 & 0.97 [0.96, 0.98] & -0.47 [-0.49, -0.45] & -0.03 [-0.03, -0.02] & 0.97 / 0.51 \\
\bottomrule
\end{tabular}}
\caption{Training-free consistency gate: change in accuracy when the answer with documents disagrees with the image-only answer ($\Delta$ vs. no gate; 95\% CI). grid\_g16\_k0 = correct document, heavily blurred image.}\label{tab:gate2}
\end{table}

\section{Contamination check}
Per-split differences for the contamination check discussed in the Results.

\begin{table}[t]
\centering\small
\resizebox{\columnwidth}{!}{%
\begin{tabular}{lccc}
\toprule
Model & $\Delta$ acc clean & $\Delta$ acc adverse & $\Delta$ follows text \\ \midrule
internvl3 & 0.01 [-0.04, 0.08] & 0.06 [-0.02, 0.13] & -0.03 [-0.13, 0.08] \\
llava05b & 0.01 [-0.05, 0.07] & 0.01 [-0.06, 0.07] & -0.02 [-0.12, 0.08] \\
llava16 & 0.01 [-0.04, 0.07] & 0.05 [-0.03, 0.12] & -0.00 [-0.11, 0.10] \\
qwen2vl & 0.00 [-0.04, 0.06] & 0.04 [-0.04, 0.12] & -0.01 [-0.12, 0.10] \\
\bottomrule
\end{tabular}}
\caption{Contamination check: difference between items from COCO train2017 (likely seen in instruction tuning) and val2017.}\label{tab:contam}
\end{table}

\begin{table*}[!t]
\centering\small
\resizebox{\textwidth}{!}{%
\begin{tabular}{lcccc}
\toprule
Model & Hugging Face id & commit & weights & wall-clock (h) \\ \midrule
internvl3 & OpenGVLab/InternVL3-8B-hf & 259a3b6 & 4-bit NF4 & 16.7 \\
llava16 & llava-hf/llava-v1.6-mistral-7b-hf & 2424fdd & 4-bit NF4 & 8.5 \\
qwen2vl & Qwen/Qwen2-VL-7B-Instruct (max 401,408 pixels) & eed1309 & 4-bit NF4 & 10.0 \\
llava05b & llava-hf/llava-interleave-qwen-0.5b-hf & 1090956 & bfloat16 & 1.4 \\
\bottomrule
\end{tabular}}
\caption{Models and runs.}\label{tab:models}
\end{table*}

\section{Validation of the cue-combination estimator}
Simulated models with planted weights (tests/test\_recovery.py). With low measurement noise the estimator recovers approximately (w\_I, w\_T) = (1.000, 0.999), (0.601, 1.798) and (1.389, 0.491) for planted (1, 1), (0.6, 1.8) and (1.4, 0.5). With high noise, over 16 independent draws, the corrected estimates averaged (0.70, 1.64) for planted (0.7, 1.6), with nominal confidence-interval coverage, while least squares averaged (0.45, 1.20). Exact values depend on the simulator's seeds; the shipped test (tests/test\_recovery.py) checks the same properties with tolerances.

\section{Prompts}
\begin{itemize}[leftmargin=*,nosep]
  \item Free-form: [documents] question + 'Answer the question using a single word or phrase.'
  \item Multiple choice / A--B: [documents] question + lettered options + 'Answer with the option's letter from the given choices directly.'
  \item POPE: 'Is there a $\langle$object$\rangle$ in the image?' + 'Answer yes or no.'
  \item Caption: 'Describe this image in one sentence.'
  \item Documents are listed as 'Retrieved documents:' followed by numbered lines.
\end{itemize}

\section{Models, compute and reproducibility}

The model configurations and run details are
summarized in Table~\ref{tab:models}.

All runs used transformers 4.57.3, PyTorch 2.8 and bitsandbytes 0.50 on one NVIDIA RTX 4090 (24 GB), with two runs sharing the GPU (about 19 hours of elapsed time in total; the individual run durations sum to 36.6 hours). Each run answered all 145,701 queries of the evaluation split without errors. The evaluation split is the seeded list `subsets/main2k.txt` (md5 d04122653ff2360bab2d02269240c79a), run with the configuration `configs/run.yaml`.

Code, the item list, per-instance predictions (provenance/, one table per model), the run configuration (configs/run.yaml) and the audit script are released with the paper.

\end{document}